\documentclass[sigconf]{acmart}
\AtBeginDocument{%
  }

\setcopyright{acmlicensed}
\acmDOI{XXXXXXX.XXXXXXX}
\acmConference[CODS'25]{CODS}{December 17--20,
  2025}{Pune, India}
\acmISBN{978-1-4503-XXXX-X/2018/06}

\usepackage{subcaption}
\usepackage{dblfloatfix}
\begin{document}

\title{V-SAT: Video Subtitle Annotation Tool}

% \affiliation{%

\author{Arpita Kundu}
\affiliation{
  \institution{LTIMindTree, India}}
  %\city{Hekla}
  %\country{India}}
\email{kunduarpita2012@gmail.com}

\author{Joyita Chakraborty}
\affiliation{
  \institution{LTIMindTree, India}}
  \email{joyita.ckra@gmail.com}

\author{Anindita Desarkar}
\affiliation{
  \institution{LTIMindTree, India}}
  \email{aninditadesarkar@gmail.com}

\author{Aritra Sen}
\affiliation{
  \institution{LTIMindTree, India}}
  \email{aritra_sen@outlook.com}

\author{Srushti Anil Patil}
\affiliation{
  \institution{LTIMindTree, India}}
  \email{srushtipatil7418@gmail.com}

%\author{Somsuvra Chatterjee}
%\affiliation{
%  \institution{LTIMindTree, India}}

%\author{Rahul Saha}
%\affiliation{
%  \institution{LTIMindTree, India}}

%\author{Ansuman Das}
%\affiliation{
%  \institution{LTIMindTree, India}}

\author{Vishwanathan Raman}
\affiliation{
  \institution{LTIMindTree, India}}  \email{vishwanathanraman@gmail.com}

%\author{Anup Karade}
%\affiliation{
%  \institution{LTIMindTree, India}}

% \affiliation{%

% \affiliation{%

% \affiliation{%

% \affiliation{%

% \affiliation{%

% \affiliation{%

\renewcommand{\shortauthors}{Kundu et al.}

\begin{abstract}
The surge of audiovisual content on streaming platforms and social media has heightened the demand for accurate and accessible subtitles. However, existing subtitle generation methods—primarily speech-based transcription or OCR-based extraction—suffer from several shortcomings, including poor synchronization, incorrect or harmful text, inconsistent formatting, inappropriate reading speeds, and the inability to adapt to dynamic audio-visual contexts. Current approaches often address isolated issues, leaving post-editing as a labor-intensive and time-consuming process. In this paper, we introduce V-SAT (Video Subtitle Annotation Tool), a unified framework that automatically detects and corrects a wide range of subtitle quality issues. By combining Large Language Models (LLMs), Vision-Language Models (VLMs), Image Processing, and Automatic Speech Recognition (ASR), V-SAT leverages contextual cues from both audio and video. Subtitle quality improved, with the SUBER score reduced from 9.6 to 3.54 after resolving all language mode issues and F1-scores of \textasciitilde 0.80 for image mode issues. Human-in-the-loop validation ensures high-quality results, providing the first comprehensive solution for robust subtitle annotation.
\end{abstract}





\maketitle

\section{Introduction}
The amount of audiovisual content has surged dramatically with the rapid growth of online platforms such as YouTube, Amazon Prime, and Netflix \footnote{https://www.insiderintelligence.com/insights/ott-video-streaming-services/}, etc. Subtitles, which are snippets of timed text displaying spoken content on screen, have become an essential feature accompanying this content. Recent surveys \footnote{https://idealinsight.co.uk/infographics/netflix-captions} confirm this growing trend: for example, 85\% of Netflix users (54\% for Amazon Prime and 37\% for Disney+) report opting to watch content with captions, and younger viewers under 30 are particularly likely to enable subtitles. Subtitles not only support viewers with hearing impairments but also benefit audiences in noisy environments, with low-quality audio, or when clarity of accents and specialized content is required. This growing demand underscores the importance of high-quality subtitle generation and presentation.

Despite these benefits, current subtitle generation and extraction methods face significant challenges that hinder user experience. Speech-based approaches \cite{papi2023direct, karakanta2021simultaneous}, which rely on transcribing audio, and OCR-based methods \cite{ramani2020automatic, yu2025eve}, which extract text from video frames, often suffer from inconsistencies. Common problems include inaccurate start and end timestamps, poor synchronization with spoken dialogue, and failure to represent non-speaker audio such as background noise, silence, or music. Generated subtitles may also contain incorrect content, including spelling errors, grammatical mistakes, mistranslations, or harmful words. Furthermore, formatting inconsistencies—such as irregular font size, color, or positioning—can reduce readability, sometimes overlapping with important on-screen visuals. Additional issues include inappropriate reading speed, measured by Characters Per Line (CPL) and Characters Per Second (CPS), and the lack of contextual adaptation to dynamic audio-visual content. These limitations collectively increase viewers’ cognitive load and disrupt immersion.

Existing literature offers a range of subtitle-generation models—some focus on speech transcription \cite{sperber2020speech}, while others use OCR for text extraction. Some works \cite{mocanu2021automatic, tapu2019deep, masiello2023synchro} attempt to solve specific problems, such as improving synchronization or optimizing subtitle positioning. Others address linguistic quality by reducing spelling and translation errors. However, these solutions are often narrow in scope, targeting one or two isolated issues rather than approaching the problem holistically \cite{gorman2021adaptive, gaido2024automatic}. Moreover, current methods rarely integrate contextual information from both audio and video, leading to subtitle texts that fail to dynamically adapt to scene changes. Subtitle post-editing today is still largely manual, involving labor-intensive corrections by human editors, which is both time-consuming and costly \cite{arroyo2024customization, may2025choices, becerra2024dialogue, koponen2020mt}. To date, there is no robust solution that simultaneously tackles the broad spectrum of challenges in subtitle generation and quality assurance.

\textbf{Key Contributions:} In this work, we propose V-SAT (Video Subtitle Annotation Tool), a unified system designed to address all major challenges in subtitle generation and post-editing. Given a raw video and subtitle file, V-SAT automatically detects inconsistencies across timing, content accuracy, formatting, positioning, and reading speed. Importantly, it not only flags these issues but also suggests automated corrections, leveraging the combined potential of Large Language Models (LLMs), Vision-Language Models (VLMs), Image Processing (IP) techniques, and Automatic Speech Recognition (ASR). 
By integrating multimodal contextual cues from both audio and video, our tool provides more accurate and adaptive subtitle generation. 
Finally, V-SAT includes an interactive human-in-the-loop validation step. In addition to automatically detecting and correcting issues, the tool also enables users to manually annotate content even when no issues are detected, thereby broadening the scope of correction. Subtitle quality was evaluated using SUBER for language mode issues and F1-score for image mode issues. The SUBER score improved from 9.6 to 3.54 after corrections, while subtitle positioning and font color issues achieved F1-scores of 0.82 and 0.80. To the best of our knowledge, this is the first comprehensive framework that attempts to jointly solve the wide range of subtitle quality issues in a single automated pipeline.

\section{Related Work}
Prior works on subtitling mainly falls into two streams-- subtitle generation \cite{papi2023direct} and subtitle extraction \cite{yu2025eve}. This categorization mainly reflects speech-related studies that derive subtitles from audio, and vision or OCR-related studies that recover on-screen text from video frames—each bringing different strengths and limitations. Below, we organize prior work along these dimensions and highlight targeted studies that address specific subtitle-quality issues (synchronization, positioning, compression), as well as gaps that motivate our approach.

Generation methods treat subtitling as Speech Translation (ST) problem \cite{sperber2020speech}, using either cascaded Automatic Speech Recognition + Machine Translation or direct end-to-end models \cite{papi2023direct, waibel1991janus}, often based on Conformer encoders \cite{gulati2020conformer}. While direct models reduce error propagation, practical pipelines still struggle with segmentation, timestamps, and translation \cite{koponen2020mt, papi2023direct}. Extraction approaches rely on OCR or vision–language models to recover on-screen text \cite{yu2025eve, ramani2020automatic, alonzo2022beyond}, but OCR often loses temporal context and vision-based models face limits in handling bounding boxes and long videos. Several open-source and commercial tools exist such as VideoOCR \footnote{https://github.com/devmaxxing/videocr-PaddleOCR} and PyVideoTrans \footnote{https://github.com/jianchang512/pyvideotrans} and other commercial cloud services like OpenVision \footnote{https://vision.aliyun.com/} and GhostCut \footnote{https://cn.jollytoday.com/home/}, though with restricted scope.

Beyond subtitle generation and extraction, specific subtitle quality issues remain. These include inaccurate timestamps \cite{masiello2023synchro}, poor synchronization under noise or multi-speaker settings \cite{may2025choices, alonzo2022beyond}, suboptimal positioning that overlaps visual content \cite{tapu2019deep, mocanu2021automatic}, and length or readability constraints in prompting-based compression methods \cite{gaido2024automatic}. Resources for evaluation are sparse \cite{karakanta2020must, guan2025trifine}, and recent LLM-based customization approaches \cite{arroyo2024customization} show mixed results.

In sum, existing methods either (i) focus on subtitle generation or extraction without integrated quality assurance or (ii) address one or two isolated subtitle quality issues. There is no comprehensive system that jointly detects the broad spectrum of subtitle issues (timing, content, format, positioning, reading speed, and contextual adaptation) and proposes corrections while remaining human-in-the-loop. This gap motivates our V-SAT tool, which aims to unify detection and correction across these dimensions using multimodal models and interactive validation.

\section{Proposed Solution Outline}
The proposed solution outline of the V-SAT tool is as follows: 

% \begin{figure}[h]
%   \centering
%   \includegraphics[width=\linewidth, height=4cm]{proposed-archt-final}
%   \caption{Proposed solution outline of the V-SAT tool}
%   %\Description{A woman and a girl in white dresses sit in an open car.}
%   \label{proposed_architecture}
% \end{figure}

\begin{itemize}

\item \textbf{Input:} The system takes a raw video file (without subtitles) and a original subtitle file in .srt or .vtt format as input. 

\item \textbf{Issues detection:} V-sat detects and corrects subtitle issues in two modes: language and image. Language mode issues include spelling and grammar errors, harmful words, timing mismatches, non-words, and segmentation problems such as overly long subtitles, etc. Image mode issues involve poor subtitle positioning and inappropriate font color in video frames. 

% Issues are then detected using LLM, ASR, and image processing techniques.

\item \textbf{Display issues and suggestion: } Detected issues are displayed with suggested corrections, timestamps, and corresponding audio content.

\item \textbf{Human Validation:} All subtitles, even those without detected issues, are open for human validation. Annotators can accept or override suggested corrections. 
%through manual annotation.

\item \textbf{Output:} Outputs are the final subtitle file similar to the uploaded format 
%(.srt/.vtt)
and the video embedded with the subtitle.

\end{itemize}

\section{System Overview}
In this section, we present an overview of V-SAT tool. 

\subsection{Pre-processing}
In this sub-section, we describe the pre-processing of input files. The system requires two inputs: (a) a raw video file without subtitles and (b) a subtitle file in .srt/.vtt format. The subtitle file is first converted into CSV, and its timestamp information is used as a reference. For each timestamp, the corresponding audio and video clips are extracted separately.

\subsection{Feature Components}
In this sub-section, we describe the issues addressed by the V-SAT tool in video subtitles and outline the approaches used for their detection and correction.

\begin{figure}[h]
  \centering
  \includegraphics[width=1\linewidth]{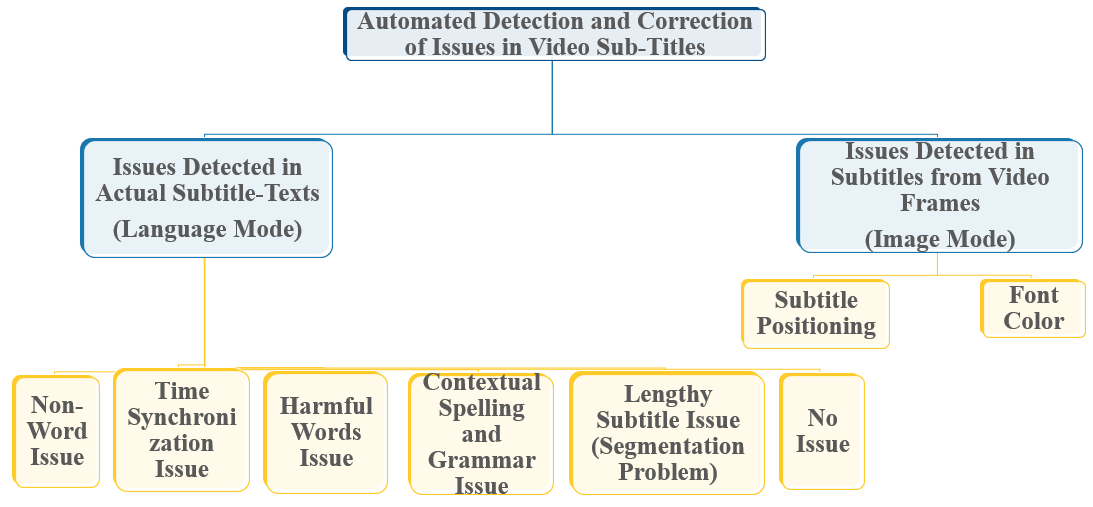}
  \caption{Summary of issues addressed in the V-SAT tool}
  %\Description{A woman and a girl in white dresses sit in an open car.}
  \label{proposed_issues}
\end{figure}

\subsubsection{Language mode issues:} In this section, we briefly discuss the issues in actual subtitle text.

\begin{itemize}
\item \textbf{Issue 1: Contextual spelling and grammar-} This issue arises when a word is spelled correctly but used incorrectly in context. For example, in a cooking video, dessert is contextually correct, whereas desert is not. To detect such cases, we prompt a Large Language Model (LLM) in zero-shot mode, using the current subtitle text along with its preceding context (subtitle texts). Unlike grammar checking, the goal is to identify contextually inappropriate words. Once identified, corrections are generated by prompting the LLM with a predefined set of rules, and each correction is validated against these rules. Overall, we find that this approach improves both detection and correction accuracy.
\begin{itemize}
\item \textit{Rule 1:} Correct words should NOT add a suffix (like -y, -ness, -ful, -less, -ed etc.) to change the grammatical category of the misspelled words.
\item \textit{Rule 2:} Correct words must be meaningful and contextual.
\item \textit{Rule 3:} Correct words differ only in spelling but fits the context better.
\end{itemize}

\item \textbf{Issue 2: Harmful words-} This issue involves filtering offensive or HAP content (hate speech, abusive language, and profanity). V-SAT uses a Large Language Model (LLM) to detect such words, which are then masked with asterisks.

\item \textbf{Issue 3: Time synchronization-} A time synchronization issue in subtitles occurs when the text on screen does not align with the corresponding audio, appearing too early, too late, or gradually drifting out of sync. In V-SAT, this issue is addressed by first extracting audio clips for each timestamp during pre-processing. Transcripts are then generated using the faster-whisper-medium model. The extracted transcripts are compared with the user-provided subtitle file using cosine similarity. If the similarity score falls below a defined threshold (i.e. 0.7), a time synchronization issue is flagged.

\item \textbf{Issue 4: Non word-} Non-word issues arise when subtitles include cues such as [music playing] or [silence] to represent background sounds instead of spoken words. In V-SAT, this is addressed by extracting audio clips for each timestamp and applying Google’s YamNet model for audio classification. YamNet predicts 521 event classes with associated probabilities, and if a class exceeds a set threshold (0.3), the corresponding event label is added to the subtitle text.

\item \textbf{Issue 5: Segmentation-} A segmentation issue occurs when subtitles are not properly divided into readable chunks, often due to lengthy text, poor line breaks, or merging multiple sentences. To detect such cases, we calculate the Characters Per Line (CPL); if it exceeds 50, a segmentation issue is flagged. The subtitle text is then split at the 50-character threshold. Since the original timestamps become inaccurate after splitting, the faster-whisper-medium model is used to generate word-level timestamps, and the last word of each split segment of subtitle texts is used to realign the timing.

\item \textbf{No issue-} Even when no issue is detected, V-SAT provides users with the option to manually annotate or edit subtitles.
\end{itemize}

\subsubsection{Image mode issues:} In this section, we briefly discuss the issues in actual subtitle text in video frames.

% \begin{figure*}[!ht]
%   \centering
%   \begin{subfigure}{0.32\textwidth}
%     \includegraphics[width=\linewidth]{segmentation.png}
%     \caption{Segmentation issue}
%   \end{subfigure}
%   \hfill
%   \begin{subfigure}{0.32\textwidth}
%     \includegraphics[width=\linewidth]{subtitle_position.png}
%     \caption{Subtitle positioning issue}
%   \end{subfigure}
%   \hfill
%   \begin{subfigure}{0.32\textwidth}
%     \includegraphics[width=\linewidth]{no_issue.png}
%     \caption{No issue}
%   \end{subfigure}
%   \caption{V-SAT Tool}
%   \label{fig:widefig}
% \end{figure*}

%####################################
% \begin{figure*}[!ht]
%   \centering
%   \begin{subfigure}{0.32\textwidth}
%     \includegraphics[height=5.8cm,width=\linewidth,keepaspectratio]{img_1.png}
%     \caption{Segmentation issue}
%   \end{subfigure}
%   \hfill
%   \begin{subfigure}{0.32\textwidth}
%     \includegraphics[height=5.8cm,width=\linewidth,keepaspectratio]{img_2.png}
%     \caption{Subtitle positioning issue}
%   \end{subfigure}
%   \hfill
%   \begin{subfigure}{0.32\textwidth}
%     \includegraphics[height=5.8cm,width=\linewidth,keepaspectratio]{img_3.png}
%     \caption{No issue}
%   \end{subfigure}
%   \caption{V-SAT Tool}
%   \label{fig:widefig}
% \end{figure*}

\begin{itemize}
\item \textbf{Issue 6: Subtitle positioning-} This issue occurs when subtitles are placed in a way that blocks important visual content or distracts the viewer. By default, subtitles appear at the bottom center, but problems arise when that area already contains text (e.g., speaker names, news tickers, or graphics) or when subtitles cover key visuals, are misaligned, or inconsistently positioned. To detect such issues, video clips are extracted per timestamp, and the first frame is analyzed. Using OpenCV, a saliency map is generated for each frame, and an overlap score is calculated between salient objects and the default subtitle position. If the score exceeds a threshold (0.006), a positioning problem is flagged. To correct it, the x and y coordinates of subtitles are shifted from the bottom center to alternative positions (e.g., middle center), and the final position is chosen where the overlap score is minimal.

\item \textbf{Issue 7: Subtitle font color-} This issue arises when subtitles are difficult to read due to poor visual design, such as inappropriate font color. Font color issues involve low contrast, distracting choices, or inconsistent usage. To address this, the subtitle background region is extracted from each frame, and its average color is converted to a brightness value. If brightness exceeds 128, the subtitle font is rendered in black; otherwise, it is rendered in white to ensure readability.
\end{itemize}

\subsubsection{Human in the loop:} User can validate the output of the tool.

%After detecting and correcting each issue in video subtitle text, we have kept the option of final validation to be performed by a user of our tool.

% \begin{figure*}[t]
%   \centering
%   \includegraphics[width=0.2\textwidth]{image_1}\\[1ex]
%   \includegraphics[width=0.2\textwidth]{image_2}\\[1ex]
%   \includegraphics[width=0.2\textwidth]{image_3}
%   \caption{An example figure spanning both columns with multiple stacked images.}
%   \label{fig:widefig}
% \end{figure*}

\subsection{Demonstration Interface}
\begin{figure}[h]
  \centering
  \includegraphics[width=\linewidth]{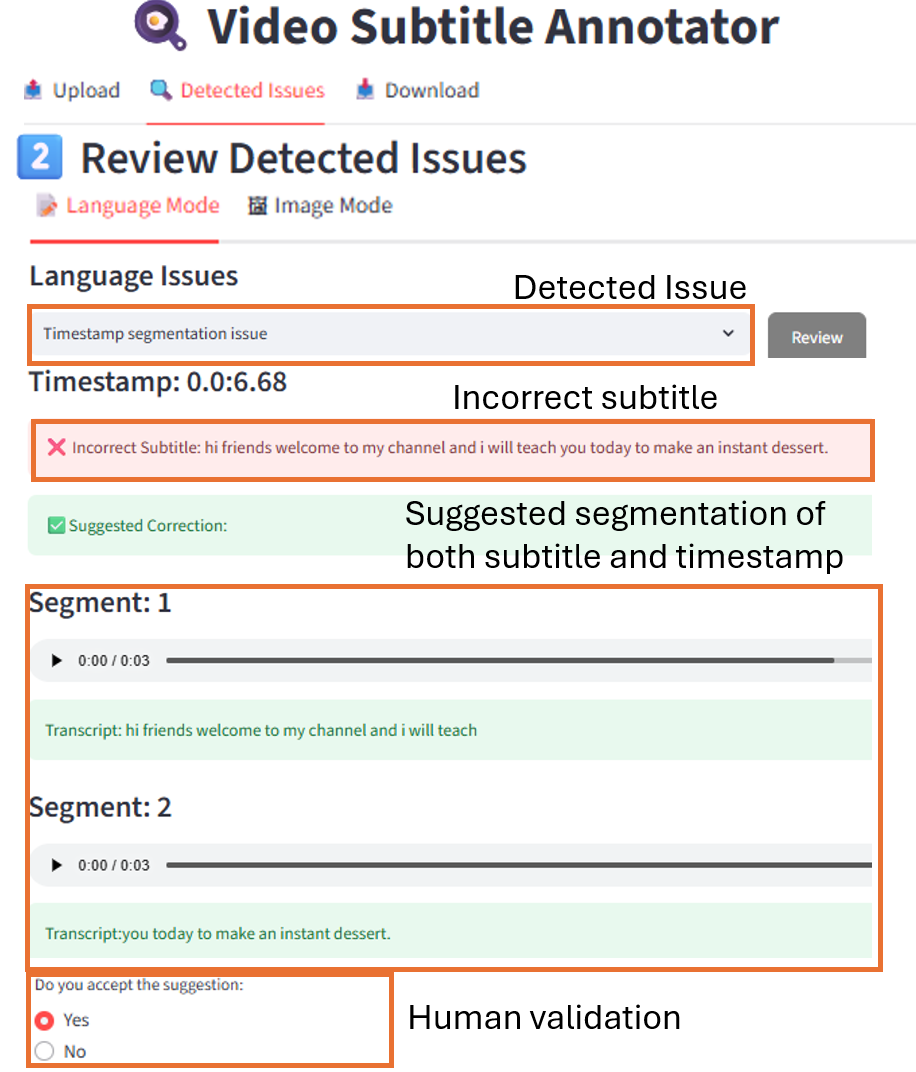}
  \caption{System interface for segmentation issues}
  \label{segmentation_issue}
\end{figure}

\begin{figure}[h]
  \centering
  \includegraphics[width=\linewidth]{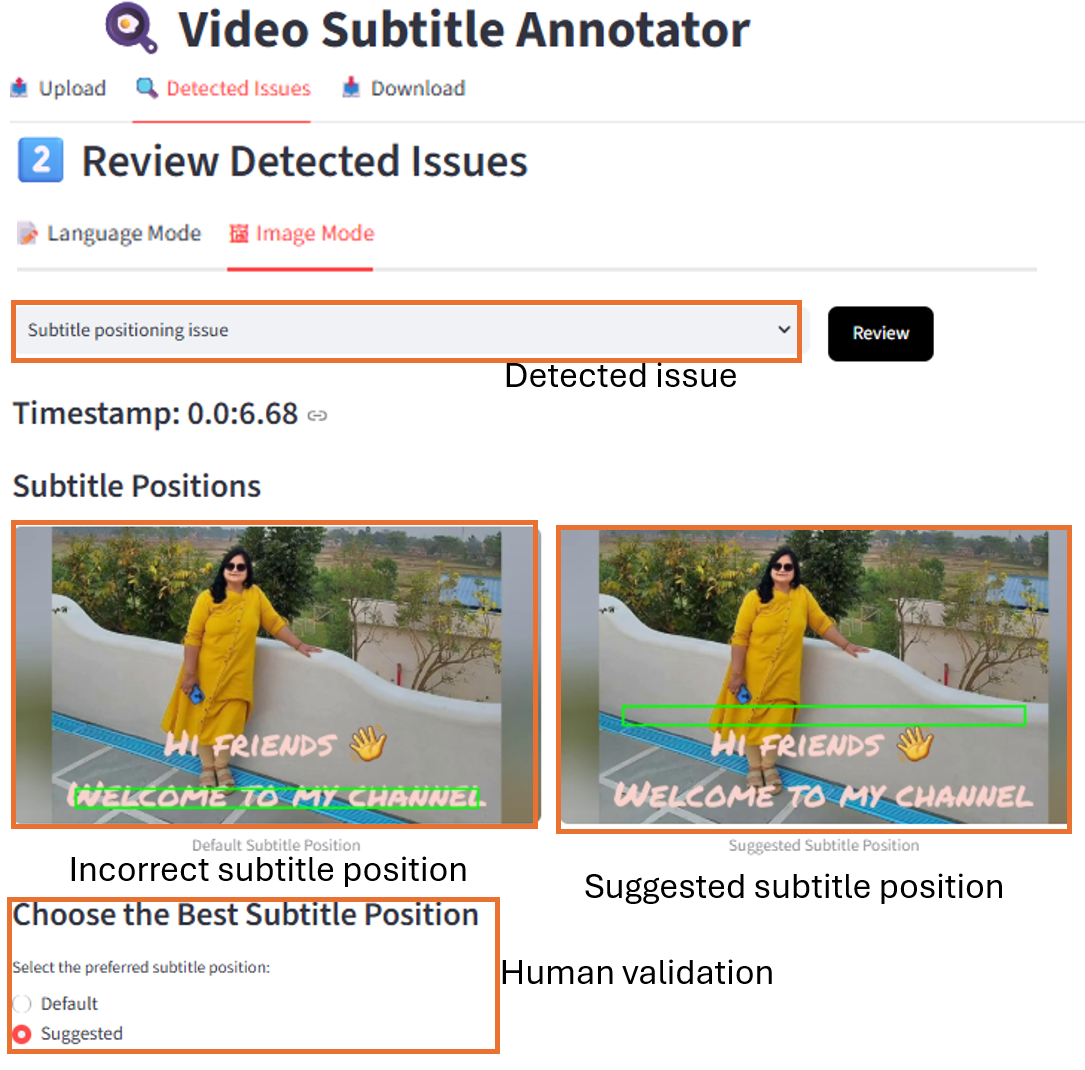}
  \caption{System interface for subtitle positioning issue}
  \label{subtitle_positioning_issue}
\end{figure}

The frontend UI, built with Streamlit, is organized into tabs. In the first tab, users can upload a raw video and a subtitle file (.srt or .vtt format). Separate tabs handle language and image-related issues in subtitle texts, with dropdowns listing specific issues. For example, see figure \ref{segmentation_issue} for segmentation issue and figure \ref{subtitle_positioning_issue} for subtitle positioning issue. The tool detects subtitle issues for each timestamp, provides suggested fixes, and lets users preview audio alongside the original and corrected text. For issues 1 to 4, users can manually edit subtitles; for others, they can accept or reject the suggestions. Users can then download the final subtitle file similar to the uploaded format (.srt/.vtt) and the video embedded with the subtitle. The demonstration video is available at \url{https://youtu.be/zBg6DnKWcuA}.

%The tool automatically detects issues, showing timestamps with erroneous subtitles, shows suggested corrections, and allows for manual direct correction by a user in the interface. Each correction can be validated by previewing the audio, erroneous text, and corrected text. For issues 1, 2, 3, and 4 a user can directly type and change the subtitle text whereas for rest of the issues a user can accept or reject the suggested correction.

%\vspace{-3mm}

%The frontend UI is developed in Streamlit, with functionalities organized into separate tabs. In the first tab, a user of the V-SAT tool can upload a raw video file along with a subtitle file in either .srt or .webvtt format. Dedicated tabs are provided for detecting language-related and image-related issues, with feature components for each issue presented in a dropdown list. A user can select any option, and the tool will automatically detect issues and display the exact timestamp along with the corresponding erroneous subtitle text. The interface also provides an option to correct the detected issues directly from the frontend. Each correction can then be cross-validated by the user: for the given timestamp, the exact audio content, erroneous subtitle text, and corrected subtitle text are all available for preview. Once validated, a final CSV file can be generated, containing timestamps and the corrected subtitle texts. Finally, the tool also provides an option to burn the corrected subtitles into the raw video.

\subsection{Tools and Techniques}
The V-SAT tool leverages multiple state-of-the-art models and libraries to detect and correct subtitle issues. For language-related issues, it employs the GPT-4o mini model as the Large Language Model (LLM). For video analysis, it uses the OpenCV library to extract and process frames. Automatic Speech Recognition (ASR) is performed using OpenAI’s Whisper model, which provides accurate transcripts along with word-level timestamps. For audio event detection, V-SAT integrates Google’s YamNet model. Together, these components enable robust multimodal analysis of subtitles across language, audio, and visual dimensions. The system requirements to run this tool includes python 3.10, 16gb ram, and i5 intel processor.

\subsection{Experimental Results}
Experiments are done on a corpus of ten videos of average 3 minute duration. Both videos and ten transcript files are downloaded from YouTube. The ground truths for these are majority voted annotations of 3 annotators.

To measure the quality of corrected subtitles after resolving each issue, two different metrics are used: the SUBER score \cite{wilken2022suber} for language mode issues and the F1-score for image mode issues. Scores are measured with respect to the ground truths. A lower SUBER score indicates better alignment with the ground truth, while a higher F1-score reflects better detection accuracy.

As shown in the table \ref{tab:suber}, input subtitle file uploaded by user with all issues had an initial SUBER score of 9.6. After gradually resolving all language-mode issues, this score was significantly reduced to 3.54. Furthermore, table \ref{tab:f1score} illustrates for issues 6 and 7, we obtained F1-scores of 0.82 and 0.80, respectively.

%\vspace{-2mm}
\begin{table}[t] % table* spans across both columns
  \centering
  \begin{minipage}{0.48\linewidth}
    \centering
    \begin{tabular}{lc}
    \toprule
    System output  & SUBER score \\
    \midrule
    Input subtitle file & 9.60\\
    Issue1 & 9.10\\
   Issue\_1+2  & 9.09\\
   Issue\_1+2+3 & 6.57\\
  Issue\_1+2+3+4 & 4.55\\
  Issue\_1+2+3+4+5 & 3.53\\
  All\_lang\_issue & 3.54\\
  \bottomrule
\end{tabular}
    \subcaption{}
    \label{tab:suber}
  \end{minipage}\hfill
  \begin{minipage}{0.48\linewidth}
    \centering
     \begin{tabular}{lc}
    \toprule
    Model  & F1-score \\
    \midrule
    Issue 6 & 0.82\\
    Issue 7 & 0.80\\
  \bottomrule
\end{tabular}
    \subcaption{}
    \label{tab:f1score}
  \end{minipage}
\caption{Performance after resolving issues in (a) language and (b) image mode which are described in section 4.2.1.}
\end{table}
\vspace{-2mm}

\section{Potential Applications}
This work has several applications. It includes real-time subtitle personalization \cite{arroyo2024customization, gorman2021adaptive}, where subtitles can be dynamically adapted to user preferences such as reading speed, text size, or display position, thereby enhancing individual viewing experiences \cite{arroyo2024customization, gorman2021adaptive}. In scenarios involving multiple speakers, the system can support selective subtitling, allowing viewers to focus on specific speakers or narrative threads to optimize video understanding. Further, this may help to  explore different aspects of a story, optimizing their video understanding experience. It also enables movie and video subtitle translation, where subtitles can be corrected and adapted across languages while maintaining synchronization and contextual fidelity \cite{yu2025eve}. Furthermore, integration into object-based media workflows allows subtitles to function as modular components, enabling structured, context-aware adaptation across diverse devices and platforms. Beyond entertainment, the framework can be applied in education, accessibility for hearing-impaired users, live events, and multilingual conferencing, providing scalable and adaptive solutions for high-quality subtitling.

\section{Conclusion}
In this paper, we introduced V-SAT (Video Subtitle Annotation Tool), a novel and comprehensive framework that integrates LLMs, VLMs, IP, and ASR to automatically detect and correct a wide range of subtitle quality issues within a single platform. Unlike prior approaches that address problems in isolation, V-SAT effectively combines temporal and contextual information from both audio and video to ensure more accurate and adaptive subtitle correction. Each automated correction is followed by an interactive human validation step, enabling a robust human-in-the-loop process that enhances reliability. Furthermore, beyond correcting the predefined issues identified in this work, V-SAT empowers users to manually annotate other problems, ensuring greater flexibility and coverage. 
%The V-SAT tool unifies automated detection, correction, and human oversight, paving the way for adaptive, accessible subtitling systems. 
Future work includes real-time correction, multilingual support, and integration with streaming platforms.


\bibliographystyle{ACM-Reference-Format}
%\bibliography{sample-base}

\appendix

\end{document}